\documentclass[conference]{IEEEtran}
\IEEEoverridecommandlockouts
\usepackage{cite}
\usepackage{amsmath,amssymb,amsfonts}
\usepackage{algorithmic}
\usepackage{graphicx}
\usepackage{textcomp}
\usepackage{booktabs}
\usepackage{xcolor}
\usepackage{multirow}
\usepackage{multicol}
\def\BibTeX{{\rm B\kern-.05em{\sc i\kern-.025em b}\kern-.08em
    T\kern-.1667em\lower.7ex\hbox{E}\kern-.125emX}}
\begin{document}    

\title{Cross-User Adaptation for sEMG Gesture Recognition via Continual Learning\\

}

\author{
\IEEEauthorblockN{Aran Ramakrishnan, Shubhi Mishra, Tyler Ham, Assel Kembay, and Jason Eshraghian}
\IEEEauthorblockA{\textit{Department of Electrical and Computer Engineering} \\
\textit{University of California, Santa Cruz, CA 95064, USA}\\
Email: \{aramakr4, smishr21, tham, akembay, jsn\}@ucsc.edu}
}

\maketitle

\begin{abstract}
Prosthetic and assistive interfaces increasingly depend on decoding hand gestures from surface electromyography (sEMG). Because the signal is shaped by person-specific factors, the recurrent neural networks (RNNs) commonly used to model these temporal streams must adapt to new wearers and new gesture classes after deployment. Continual
learning (CL) enables this adaptation without retraining from scratch, yet applying it to recurrent sEMG models is not straightforward. Standard recurrent training practices can destabilize learning, and which CL strategy holds up under realistic cross-user drift is unestablished. We evaluate three representative CL strategies under \textit{domain}- and \textit{class}-incremental drift on NinaPro DB1, and find experience replay to outperform elastic weight consolidation (EWC) and naive fine-tuning. It reaches 83.3\% final accuracy for class-incremental drift while limiting forgetting to 1.8-5.8\% accuracy drop, as against 16.2-28.5\% for EWC and naive training. We further show these models must be trained on shuffled, independently-windowed segments. Streaming windows in temporal order collapses the network into predicting a single dominant gesture, while a shuffled, stateless regime trains stably. Finally, hand-crafted time-domain features raise per-task accuracy by up to $\sim$40\% over raw signals, but only replay preserves that gain under class-incremental drift.
\end{abstract}


\section{Introduction}
Gesture recognition from surface electromyography (sEMG) has gained particular significance with the growing demand for precise and reliable control interfaces in prosthetic and assistive technologies. Recognition is person-specific. Muscle anatomy, electrode placement, and movement style all influence the signal, so a model narrowed to a single wearer typically outperforms a general-purpose decoder. However, personalization requires regimes that adapt to new wearers and expanding gesture vocabularies. Continual Learning (CL) provides exactly this capability, updating a deployed model on new subjects and gestures without retraining from scratch. The primary challenge in CL is catastrophic forgetting, where learning new tasks causes accuracy to decline on past ones~\cite{Kirkpatrick_2017}. CL methods to counter forgetting are maturing rapidly, but they are developed and benchmarked almost entirely outside the recurrent models that dominate sEMG processing, leaving limited insight on which CL strategies survive realistic cross-user drift, or whether standard recurrent training practices even hold up under it.

We evaluate naive, EWC, and experience-replay strategies across domain-incremental (DIL) and class-incremental (CIL) formulations, using both classic recurrent networks and modern state-space models~\cite{gu2024mambalineartimesequencemodeling} on NinaPro DB1~\cite{atzori2014electromyography,atzori2015characterization}. Our analysis yields three findings that build on one another. First, comparing CL strategies across both scenarios and both backbones, we find experience replay the most robust, and the choice of CL method outweighs the choice of recurrent architecture. This prompted us to focus on memory retention strategies, leading to the second finding. Carrying hidden state across tasks is ill-posed, and naively training in temporal order collapses the network to a majority-class predictor. This is an effect we trace primarily to batch ordering rather than to recurrent memory. We demonstrate a solution by purging the statefulness between tasks, and also shuffling data samples. This corrected regime is what makes our third finding hold. Time-domain features improve how well each task is learned, but only rehearsal preserves that improvement under class-incremental drift.

\section{Prior Work} 
Most prior work on CL for sEMG generates empirical insight into how forgetting-mitigation strategies behave on this signal \cite{ChenEMG,yoneda2026generativeversusdiscriminativeapproaches,KanogaCL}. Chen et al.~\cite{ChenEMG} establish the feasibility of cross-subject adaptation, though by aggregating subjects into a few grouped sets rather than treating each as an independent sequential task. Others examine the choice of memory mechanism or model families: Kanoga et al.~\cite{KanogaCL} apply deep generative replay under class-incremental drift, and Yoneda et al.~\cite{yoneda2026generativeversusdiscriminativeapproaches} compare generative against discriminative model families under the same settings. These studies, however, evaluate predominantly within a single CL scenario and over feedforward or generative decoders, leaving the recurrent models standard for sEMG largely untested.

We build on this body of work by testing more systematically across both \textit{domain-} and \textit{class-}incremental drift, and across two recurrent families: the classic LSTM~\cite{hochreiter1997lstm} and HGRN~\cite{qin2023hgrn}, a modern state-space model (SSM) representative of current state-of-the-art sequence architectures. This systematic comparison is what leads us to improve retention by altering how recurrent state is handled across windows.

\section{Methodology}
\subsection{Dataset and Task Formulation} \label{subsec:data_task_form}
We evaluated on NinaPro Database 1~\cite{atzori2014electromyography,atzori2015characterization}, which contains sEMG data from 27 subjects, which contains three distinct exercises, each containing a set of isotonic and isometric hand and wrist gestures. Exercise 1 includes 12 gestures, Exercise 2 has 17, and Exercise 3 has 23.
We split the task into two continual learning scenarios as outlined in van de Ven \& Tolias~\cite{vandeven2019three}:

\textbf{Domain-Incremental Learning (DIL):} In this setup, the gesture set is statically fixed to the gestures from Exercise 1. The model is trained sequentially across the 27 subjects, where each subject is treated as an independent continual learning task. This tests generalization across anatomical variances and sensor placement discrepancies between individuals without forgetting the core temporal patterns.

\textbf{Class-Incremental Learning (CIL):} In this setup, the model is trained sequentially across an expanding stream of data (Exercise 1 $\rightarrow$ Exercise 2 $\rightarrow$ Exercise 3). The ground truth labels for the new gestures in subsequent exercises must be offset sequentially in order to allow for CIL. For the HGRN\cite{qin2023hgrn} outlined in \S\ref{HGRN_info}, the classification head is sized statically at initialization to accommodate all classes, with the ``rest'' state (class 0) remaining shared throughout all tasks. The expansion of the class set makes this task more difficult than DIL, since the model must generalize to new labels, while retaining performance on past samples.

\subsection{Signal Preprocessing and Feature Extraction}
\label{preprocessing}

 Data was partitioned at the repetition level to ensure that no samples from the same gesture execution appeared in both sets, with a 70/30 train/test split. Following this split, the signals were normalised to achieve zero mean and unit variance, with the scaler fitted on training data and applied to both sets. Signals were then segmented using a sliding window approach~\cite{englehart2003robust}. The ground truth label for each window was determined by the active stimulus at the final sample.
\textbf{Time-Domain Feature Extraction:} Each 200-sample window was divided into $K = 10$ equal sub-windows of 20 samples (10 ms) each and transformed into a time-domain (TD) feature sequence following the Hudgins feature set~\cite{hudgins1993new} (MAV, RMS, WL, ZC, SSC) with a noise-rejection threshold $\epsilon = 0.01\sigma$ applied to ZC and SSC to suppress spurious crossings~\cite{phinyomark2012feature}. A second standardisation pass was applied to the TD feature matrix, using per-feature mean and variance fitted on training windows only.





\subsection{Base Network Architectures}
We evaluate two recurrent baseline architectures with a common configuration: $L = 4$ layers, hidden dimension $D = 128$, and a shared classification head. 

\subsubsection{Long Short-Term Memory}
The LSTM~\cite{hochreiter1997lstm} mitigates the vanishing gradient problem through explicit memory cells, making it a standard sequential baseline for this experiment.

\label{HGRN_info}
\subsubsection{Hierarchical Gated Recurrent Network}
The HGRN~\cite{qin2023hgrn} is a gated linear recurrence model that extends the standard forget-gate formulation with a hierarchical lower bound ($\gamma \in (0, 1)^D$), initialized proportionally to layer depth ($\gamma_{\text{init}} = l/L$). This biases shallow layers to hold short-term relations and deep layers to have long-range retention. The recurrence is executed using a hardware-aware chunked linear-recurrence kernel (FLA/Triton~\cite{yang2024fla,yang2024gla}).

\textbf{Shared Classification Head.} For both architectures, the hidden state at the final timestep is passed to a linear head mapping to the active class set. Under the TD feature regime, the input shape is $[N, K, 50]$ ($K=10$); under the raw regime, it is $[N, T, C]$ ($T=200$, $C=10$).

\subsection{Continual Learning Methods}

\label{stateful_stateless_comp}
\subsubsection{Naive Stateless and Stateful Baselines}

We evaluated two naive training regimes that serve as lower bounds before any memory retention mechanism is applied. In the Stateless baseline, the recurrent hidden state is reset to zero ($h_0 = \mathbf{0}$) at the start of every batch. This is the intuitive choice for independently-windowed EMG data. Consecutive windows are extracted with overlap from a fixed-length recording and carry no causal dependency across gesture boundaries. In the Stateful baseline, the hidden state is carried across batches within and across tasks, simulating an unbroken temporal stream. 

\subsubsection{Standard Experience Replay}
We implemented a rehearsal-based experience replay mechanism~\cite{rolnick2019experience} with a global reservoir-sampling memory buffer~\cite{vitter1985random} (capped 10,000 windows). At each training step, a mini-batch of $N_r = 16$ sequences is uniformly sampled from the buffer and concatenated with the current live batch into a single joint mini-batch. Replay sequences are augmented with additive Gaussian noise:

\[
\tilde{X}_{\text{replay}} = X_{\text{replay}} + \mathcal{N}(0, 0.01^2)
\]

The replay loss contribution is scaled by a weighting factor $\lambda_{\text{replay}} = 0.5$:

\[
\mathcal{L} = \mathcal{L}_{\text{CE}}^{\text{live}} + \lambda_{\text{replay}} \mathcal{L}_{\text{CE}}^{\text{replay}}
\]

\subsubsection{Online Elastic Weight Consolidation (EWC)}
As a regularisation-based alternative to rehearsal, we implemented online EWC~\cite{Kirkpatrick_2017,schwarz2018progress} for both architectures. At the boundary of each task, the empirical Fisher information $\hat{F}_i$ is estimated from the predicted class probabilities over the completed task's training data and accumulated across all prior tasks. The objective adds a quadratic penalty on deviations from the most recently completed task's weights $\theta^*$:

\[
\mathcal{L} = \mathcal{L}_{\text{CE}} + \lambda_{\text{ewc}} \sum_i \hat{F}_i \left(\theta_i - \theta^*_i\right)^2
\]

The Fisher is computed in stateless mode to remain consistent with the inference regime. We set $\lambda_{\text{ewc}}=30$, the largest value preserving plasticity across the full sequence (see~\S\ref{ewc_reg_qual}).

\subsection{Evaluation Metrics}

All metrics are computed from $R \in \mathbb{R}^{T \times T}$, where $R_{i,j}$ is accuracy on Task $j$ after training on Task $i$.



\paragraph{Final Accuracy (FA)}
Accuracy on all tasks after the final task :
\begin{equation}
    \text{FA} = \frac{1}{T} \sum_{i=1}^{T} R_{T,i}.
\end{equation}


\paragraph{Forgetting (Frgt)}
Average drop in accuracy per task after all subsequent tasks have been trained:
\begin{equation}
    F_i = \max(0, R_{i,i} - R_{T,i}),
\end{equation}
\begin{equation}
    \text{Frgt} = \frac{1}{T-1} \sum_{i=1}^{T-1} F_i.
\end{equation}


\paragraph{Plasticity}
\label{plast_def}
We measure plasticity as the mean final-epoch training accuracy over the last $\min(10, T)$ tasks to distinguish genuine learning from weight freezing:
\begin{equation}
\begin{split}
\mathrm{Plast} &= \frac{1}{|\mathcal{S}|}\sum_{t \in \mathcal{S}} a_t^{\text{train}} \\
\mathcal{S} &= \{\, T-\min(10,T)+1,\,\dots,\,T \,\}
\end{split}
\end{equation}




\section{Results}

\subsection{CL Method Comparison}
Across all evaluated configurations, replay-based methods consistently demonstrated the strongest resistance to catastrophic forgetting (Table~\ref{tab:results_single}), as reflected by their lower forgetting values. This trend held regardless of the continual-learning scenario (DIL or CIL) or the underlying model architecture (HGRN or LSTM), indicating that rehearsal is a robust strategy for stabilizing EMG-based learning. Stateless replay yielded the highest final accuracy, with CIL-LSTM reaching $83.3\%$ accuracy and CIL-HGRN reaching $81.6\%$ accuracy, while maintaining forgetting values as low as $1.8\%$--$5.8\%$ accuracy drop. EWC exhibited substantially lower final accuracies and larger forgetting ($16.2\%$--$28.5\%$) accuracy drop, and naive models also had similar forgetting measures.

\begin{table}[!ht]
\centering
\caption{Continual-learning method comparison (TD features, stateless).}
\label{tab:results_single}
\begin{minipage}{\linewidth} 
\centering
\setlength{\tabcolsep}{3.5pt}
\footnotesize
\begin{tabular}{@{}ll cc cc@{}}
\toprule
 & & \multicolumn{2}{c}{\textbf{DIL}} & \multicolumn{2}{c}{\textbf{CIL}} \\
\cmidrule(lr){3-4}\cmidrule(lr){5-6}
\textbf{Arch} & \textbf{Method} & F-Acc\,$\uparrow$ & Frgt\,$\downarrow$ & F-Acc\,$\uparrow$ & Frgt\,$\downarrow$ \\
\midrule
\multirow{3}{*}{HGRN}
 & naive  & $0.644{\scriptstyle\pm.03}$            & $0.176$          & $0.659{\scriptstyle\pm.09}$            & $0.239$ \\
 & replay & $\mathbf{0.763}{\scriptstyle\pm.01}$   & $\mathbf{0.043}$ & $\mathbf{0.816}{\scriptstyle\pm.06}$            & $\mathbf{0.018}$ \\
 & EWC    & $0.641{\scriptstyle\pm.05}$            & $0.168$          & $0.682{\scriptstyle\pm.08}$            & $0.220$ \\
\midrule
\multirow{3}{*}{LSTM}
 & naive  & $0.627{\scriptstyle\pm.02}$            & $0.212$          & $0.679{\scriptstyle\pm.05}$            & $0.245$ \\
 & replay & $\textbf{0.757}{\scriptstyle\pm.01}$            & $\textbf{0.058}$          & $\mathbf{0.833}{\scriptstyle\pm.07}$   & $\textbf{0.023}$ \\
 & EWC    & $0.681{\scriptstyle\pm.01}$            & $0.162$          & $0.679{\scriptstyle\pm.07}$            & $0.285$ \\
\bottomrule
\end{tabular}

\vspace{4pt} 
\raggedright 
\footnotesize \textit{Note:} Mean\,$\pm$\,std over $n$ runs ($n{=}3$ seeds for DIL; $n{=}10$ subjects for CIL); architecture is fixed per configuration. Best per architecture in \textbf{bold}. All stateless configurations remain plastic.
\end{minipage}
\end{table}

Despite CIL being the harder task, replay under CIL achieves lower forgetting ($1.8\%$–$2.3\%$ accuracy drop) than under DIL ($4.3\%$–$5.8\%$ accuracy drop). This can likely be attributed to the size of task streams. The CIL stream comprises only three tasks, against the 27 tasks of the DIL stream. Therefore, forgetting, which is averaged over transitions and accumulates with sequence length, is expected to be smaller under CIL on this basis alone. 
\begin{figure}[!ht]
    \centering
    \includegraphics[width=\columnwidth]{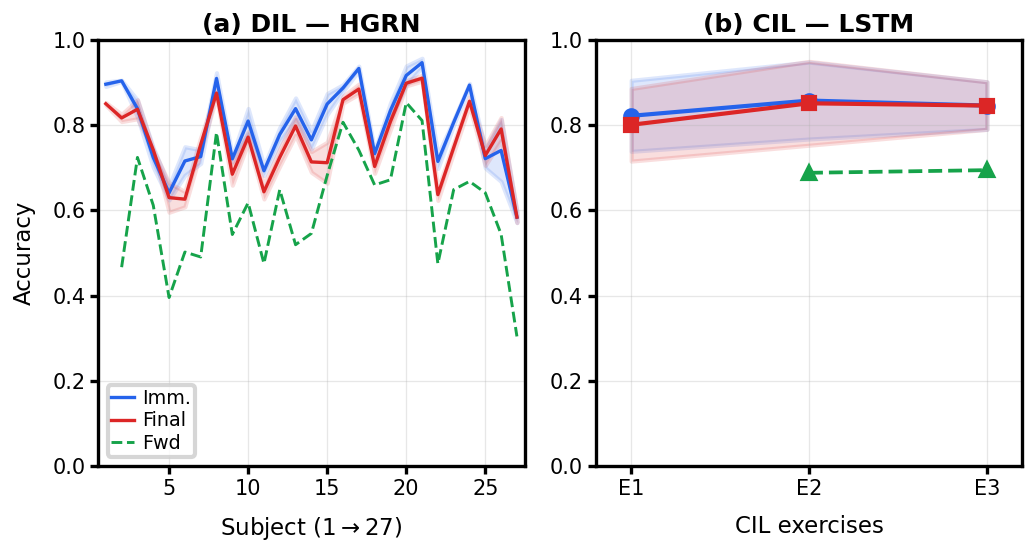}
    \caption{Task-level accuracy for replay under the DIL HGRN (a) and CIL LSTM (b) settings. Immediate (blue) and Final (red) accuracies remain closely aligned, indicating reduced forgetting. Shaded regions denote $\pm$1 standard deviation over multiple runs (seeds or subjects).}
    \label{fig:dil_replay_accuracy}
\end{figure}

\subsection{Feature Representation}
We evaluated both raw sEMG signals and time-domain (TD) Hudgins features as input representations, reflected in Table \ref{tab:rawtd}. In the DIL setting, TD features consistently produced higher accuracies, by $\sim$40\%. They also remove the training instability seen with raw windows. Whereas raw features yield high seed variance for the weaker configurations (e.g.\ LSTM naive, $\pm0.093$; LSTM EWC, $\pm0.067$), every TD DIL configuration converges to a tight range ($\text{std}\le0.03$).

However, this benefit was scenario-dependent, as in the class-incremental settings, TD features only help replay, while for naive and EWC the change was negligible. Although TD raises immediate accuracy, it also causes an increase in forgetting in non-rehearsal methods (e.g.\ LSTM EWC, $12.3\%\!\to\!28.5\%$ drop in accuracy on old tasks), causing the initial gain to be undone. Ultimately, TD features improve learning, and rehearsal is able to convert that improvement into retention. 

\begin{table}[!ht]
\centering
\caption{Raw vs.\ TD features (stateless, plastic EWC $\lambda{=}30$).}
\label{tab:rawtd}
\setlength{\tabcolsep}{4pt}
\begin{tabular}{ll cc@{\hskip 8pt} cc}
\toprule
& & \multicolumn{2}{c}{DIL} & \multicolumn{2}{c}{CIL} \\
\cmidrule(lr){3-4}\cmidrule(lr){5-6}
Arch & Method & Raw & $\Delta$TD & Raw & $\Delta$TD \\
\midrule
\multicolumn{6}{l}{\emph{Final accuracy} ($\uparrow$)} \\
HGRN & naive  & 0.544 & $+0.101$ & 0.659 & $-0.000$ \\
HGRN & EWC    & 0.570 & $+0.072$ & 0.689 & $-0.007$ \\
LSTM & naive  & 0.487 & $+0.140$ & 0.688 & $-0.009$ \\
LSTM & replay & 0.718 & $+0.039$ & 0.772 & $\mathbf{+0.061}$ \\
LSTM & EWC    & 0.485 & $+0.196$ & 0.676 & $+0.003$ \\
\midrule
\multicolumn{6}{l}{\emph{Forgetting} ($\downarrow$; $\Delta$TD${>}0$ = less forgetting)} \\
HGRN & naive  & 0.235 & $+0.059$ & 0.222 & $-0.017$ \\
HGRN & replay & 0.106 & $+0.063$ & 0.048 & $+0.030$ \\
HGRN & EWC    & 0.204 & $+0.036$ & 0.170 & $-0.050$ \\
LSTM & naive  & 0.301 & $+0.089$ & 0.165 & $-0.080$ \\
LSTM & replay & 0.053 & $-0.005$ & 0.019 & $-0.003$ \\
LSTM & EWC    & 0.305 & $+0.143$ & 0.123 & $-0.162$ \\
\bottomrule
\end{tabular}
\end{table}
\subsection{Carrying State}
\label{sec:negative-learning}

The windowing in \S\ref{preprocessing} considers each segment to be independent samples, so recurrent models should read each window with a zero state. Therefore, we see failures in two ways depending on batch ordering when trying to use this data as a continuous stream.

Considering the case where windows are presented in a temporal order, although carried state is logical, training collapses. This is because in rest-dominated sets of sEMG data the gradient is consumed by the rest class causing the network to fall into a constant predictor. If the models are shuffled, then holding the state results in irrelevant noise being injected to a batch's initial state. This is no different from a stateless regime with a random start state.

\begin{figure}[!ht]
  \centering
  \includegraphics[width=\columnwidth]{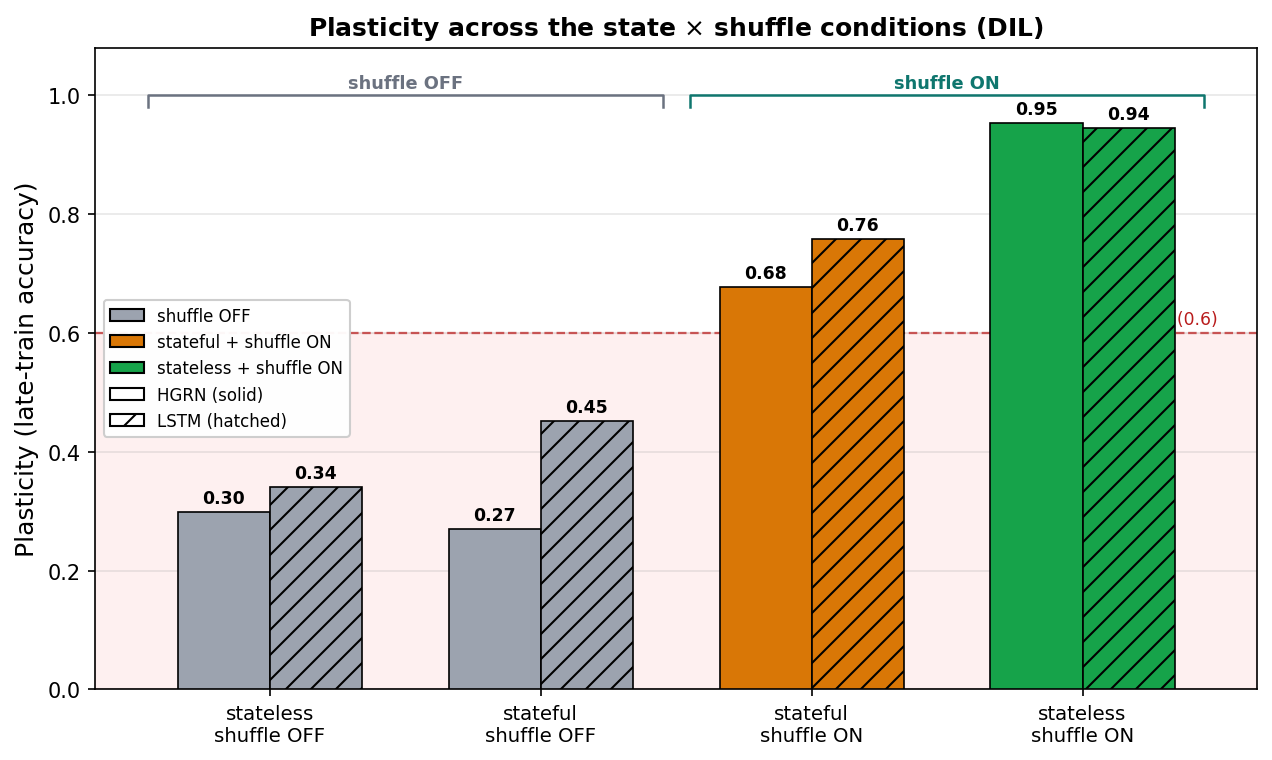}
  \caption{Plasticity (late-train accuracy) across the state$\times$shuffle conditions, DIL. Everything below the red line at $60\%$ denotes a frozen model. Both shuffle-off conditions collapse below the frozen threshold; with shuffling, stateful sits below stateless by the cost of its now-irrelevant carried state.}
  \label{fig:shuffle_state}
\end{figure}

This is confirmed by a  $2\times2$ ablation over \{stateful, stateless\} $\times$ \{shuffle on, off\} (Fig.~\ref{fig:shuffle_state}). Both shuffle-off conditions result in a plasticity collapse ($27\%$--$45\%$, below $60\%$ threshold)  into the frozen region. We even see that stateless+shuffle-off freezes, which shows that ordering drives the heavy collapse. With shuffling restored, stateful recovers to $68\%$--$76\%$ but remains below stateless at $94\%$--$95\%$, for reasons discussed above.

\subsection{EWC Regularization Quality}
\label{ewc_reg_qual}
EWC's limited effectiveness in our setting stems from two compounding factors. First, calibrating $\lambda_{\text{ewc}}$ is non-trivial at scale: values tuned on a 6-subject proxy do not transfer to the full 27-subject sweep, consistent with the theoretical finding that a fixed $\lambda$ becomes increasingly rigid as task count grows~\cite{kim2026nonequilibrium}. For the TD feature regime, $\lambda_{\text{ewc}} = 30$ was the largest value preserving plasticity across the full sequence. Second, and more fundamentally, an excessively large $\lambda$ freezes the network, producing high apparent accuracy through weight immobilization rather than genuine consolidation~\cite{kutalev2021stabilizing}. We identify frozen networks via the late-train plasticity metric (plasticity $< 0.60$ on final tasks). EWC configurations that avoid freezing still exhibit forgetting in the range of $0.162$--$0.285$  accuracy drop, confirming that Fisher-based regularization is insufficient to constrain parameter drift under the strong distributional shifts present in both DIL and CIL.

\subsection{Effect of Recurrent Architecture}
LSTM and HGRN perform comparably across CL methods and scenarios, with no configuration showing a consistent architectural advantage. The top-ranked configuration uses LSTM-CIL replay ($83.3\%$ accuracy), while the second uses HGRN-CIL replay ($81.6\%$ accuracy), a difference within one standard deviation. 

We measured architecture-dependent but hardware-independent statistics(Params and FLOPs) to gauge model performance (Table~\ref{tab:compute_cost}). The fixed buffer size for replay was about 20.18\,MB, the same across all replay configurations.

\begin{table}[!ht]
\centering
\caption{Model cost and memory footprint (TD features).}
\label{tab:compute_cost}
\setlength{\tabcolsep}{4pt}
\footnotesize
\begin{tabular}{@{}l cc cc@{}}
\toprule
 & \multicolumn{2}{c}{\textbf{DIL}} & \multicolumn{2}{c}{\textbf{CIL}} \\
\cmidrule(lr){2-3}\cmidrule(lr){4-5}
\textbf{Arch} & Params & FLOPs/win & Params & FLOPs/win \\
\midrule
HGRN & 339K & 6.68M & 345K & 6.70M \\
LSTM & 538K & 10.62M & 543K & 10.63M \\
\bottomrule
\end{tabular}

\end{table}
The HGRN model yielded about the same accuracy at 1.6x lower cost. 

\section{Conclusion}
We evaluated continual-learning strategies for sEMG-based gesture classification across 27 subjects and three exercises under DIL and CIL formulations. Experience replay outperformed EWC and naive baselines, achieving 83.3\% final accuracy with forgetting as low as 1.8\% drop in accuracy. It is clear that the dominant factor in long-term retention was the CL method rather than the recurrent architecture. Concerning stateful collapse, a 2×2 ablation attributes the collapse to batch ordering rather than recurrent state. We attribute this to windowing creating independent segments of sEMG data. Finally, time-domain features improved per-task learning but required rehearsal to retain that gain under class-incremental drift. Although we observed lower forgetting under CIL than DIL using replay, this could be attributed to the CIL stream being much smaller than the DIL stream (3 vs 27). 

These results offer a practical foundation for prosthetic control systems that must adapt to new users and gestures over time. While this paper explores the effects of different CL methods and recurrent architectures, the experiments were conducted offline, only using NinaPro DB1. Future work should examine whether results for the aforementioned CL methods and architectures transfer to other datasets in different experimental settings.

\bibliographystyle{IEEEtran}
\bibliography{references}

\end{document}